\documentclass{article}

\usepackage{microtype}
\usepackage{graphicx}
\usepackage{subcaption}
\usepackage{booktabs}
\usepackage{hyperref}

\usepackage[accepted]{icml2026}

\usepackage{amsmath}
\usepackage{amssymb}
\usepackage{mathtools}
\usepackage{amsthm}
\usepackage[capitalize,noabbrev]{cleveref}

\icmltitlerunning{DA-RAC: Distance-Aware Calibration of LLM Judges}

\usepackage{graphicx}

\begin{document}

\twocolumn[
  \icmltitle{DA-RAC: Distance-Aware Calibration of LLM Judges \\for Trustworthy AI Auditing}

  \icmlsetsymbol{equal}{*}

  \begin{icmlauthorlist}
    \icmlauthor{Cheng Wu}{equal,yyy}
    \icmlauthor{Vishal Anand}{equal,yyy}
    \icmlauthor{Jaya Krishna Mandivarapu}{yyy}
    \icmlauthor{Xiya Liu}{yyy}
    \icmlauthor{Rui Zhuang}{yyy}
  \end{icmlauthorlist}

  \center{\href{https://distance-aware-calibration.github.io/icml}{https://distance-aware-calibration.github.io/icml}}

  \icmlaffiliation{yyy}{Microsoft, Redmond, Washington, USA}

  \icmlcorrespondingauthor{Cheng Wu}{wucheng@microsoft.com}
  \icmlcorrespondingauthor{Vishal Anand}{vishal.anand@microsoft.com}

  \icmlkeywords{LLM evaluation, LLM-as-a-judge, calibration, retrieval-augmented evaluation, AI auditing, human-centered AI, interpretive technologies}

  \vskip 0.3in
]

\printAffiliationsAndNotice{\icmlEqualContribution}

\begin{abstract}
Generative AI systems are increasingly producing real-world artifacts, however their efficacy and validity are often evaluated via context-free LLM-scoring. These judges can be miscalibrated by irrelevant in-context reference examples, creating false confidence and allowing low-quality or harmful outputs to pass evaluation. We study this failure mode as context-induced miscalibration and introduce DA-RAC, a distance-aware reference-anchored calibration method for LLM judges. DA-RAC retrieves semantically and structurally similar labeled anchors for each judgement scenario, weights them by distance, and exposes neighborhood difficulty as a calibration and triage signal. On multi-run LLM-judge evaluation benchmarks, it improves calibration and reduces false-pass risk relative to zero-shot, chain-of-thought evaluation, and static-anchor baselines. Mechanistic analysis shows that judge scores vary systematically with anchor distance, while static references can induce misleading decision boundaries. Thus LLM-judgement requires not only better models, but also calibrated, auditable reference selection, especially when automated evaluation is used to support high-impact AI generated artifacts. Judgments should be grounded in relevant, inspectable, and contestable interpretive artifacts.
\end{abstract}

\section{Introduction}
\label{sec:introduction}

Generative AI systems increasingly produce cultural artifacts: stories, explanations, summaries, advice, public narratives, critiques, and design proposals.
Evaluating such artifacts goes above and beyond checking for correctness or harm avoidance.
This often depends on genre, audience, community context, historical precedent, aesthetic stance, and the values a system is meant to support~\cite{geertz1973interpretation,hall1980encoding}. A poem, a museum label, a community-facing explanation, or a public-health message may succeed in one interpretive context and be unsuccessful in another.

\begin{figure}[t]
	\centering
	\includegraphics[width=\linewidth,trim={0.85cm 10.2cm 20.45cm 0.58cm},clip]{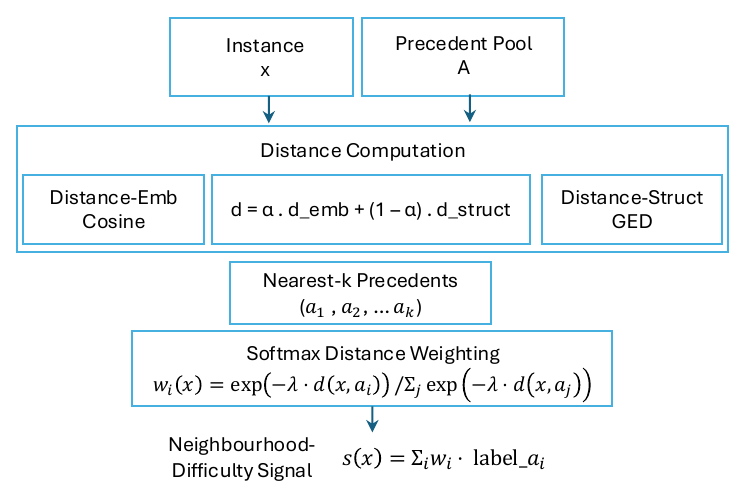}
	\caption{DA-RAC workflow: (1) input target and precedent pool; (2) compute hybrid distance combining instruction embeddings~\cite{reimers2019sentencebert} and graph edit distance~\cite{sanfeliu1983ged}; (3) select $k$ nearest precedents; (4) weight by softmax distances; (5) emit a judgement together with neighbourhood-difficulty signal.}
	\label{fig:workflow}
	\vspace{-1.0em}
\end{figure}

\paragraph{LLM as judges.}
LLM judges are often treated as scalable substitutes for human evaluation \cite{tan2025judgebench}, and in case of cultural domains they are increasingly being utilized as an \emph{interpretive technology}: systems that make judgments about meaning, relevance, quality, and value of an artifact — often influenced by prompts, examples, or learned implicit norms ~\cite{liu2023geval,saito2023verbosity,liu2023calibrating}.
In few-shot evaluation, reference examples act like \emph{precedents} which they tell the judge what kind of interpretation is appropriate~\cite{gadamer1975truth,dourish2004context}. Calibrating reference selection therefore matters as much as calibrating the judge itself.

\paragraph{Context-induced miscalibration.}
We identify a failure mode we call \emph{context-induced miscalibration}.
When an LLM judge is conditioned on irrelevant or mismatched reference examples, its decision boundary shifts.
In ordinary benchmark settings this appears as degraded accuracy or calibration~\cite{guo2017calibration}: random precedents achieve 52\% accuracy which is eighteen points \emph{below} the zero-shot baseline (Table~\ref{tab:exp1}).
Thus, an LLM judge may reward generic fluency over genre-specific success, impose dominant norms on niche artifacts, or misread a work because the supplied precedents come from the wrong interpretive background~\cite{suchman1987plans,costanza2020design}.
Existing retrieval-augmented judge methods~\cite{hasanbeig2023allure,bai2023benchmarking}may optimise accuracy rather than directly targeting the relevance of interpretive precedents.

\section{Contributions}

We identify \emph{context-induced miscalibration}, a failure mode in which irrelevant or mismatched precedents distort the judge's interpretive frame, where reference context determines whether an artifact is read according to the right genre, audience, community, or value.

We propose \textbf{DA-RAC} (Distance-Aware Reference-Anchored Calibration), a distance-aware reference anchoring method for LLM judges. DA-RAC retrieves semantically and structurally similar labelled precedents for each judgment scenario, weights them by distance, and exposes neighbourhood difficulty as a signal for human review. Technically, DA-RAC improves calibration by replacing arbitrary few-shot context with distance-aware precedents. Conceptually, DA-RAC reframes evaluation as \emph{interpretive anchoring}: grounding a judgment in relevant examples while making the basis of that judgment inspectable.

This work's interpretation is not that AI systems should replace critics, artists, communities, or domain experts. Rather, evaluation should support contextual sensitivity and human agency.
A useful evaluator should show which precedents shaped its judgment, indicate when a case lies far from known examples, and route contested cases back to humans.

\section{Method}
\label{sec:method}

DA-RAC treats few-shot examples not merely as prompt-engineering artifacts but as \emph{interpretive precedents}. In cultural evaluation, precedents establish which genres, audiences, and criteria are relevant to a judgment. A mismatched precedent can shift the evaluator toward the wrong interpretive frame, while a relevant precedent can help the judge situate the target artifact.

\subsection{Problem Formulation}

Given pairwise evaluation instances $\mathcal{T} = \{(x_i, a_i, b_i, y_i)\}_{i=1}^N$ partitioned into a precedent pool $\mathcal{A}$ (70\%) and an evaluation set $\mathcal{E}$ (30\%), DA-RAC replaces uniform few-shot with:
\[
\hat{y} = f_\theta(x | \mathcal{P}(x, \mathcal{A}_k(x), w(x)))
\]
where $\mathcal{A}_k(x)$ are the $k$ nearest precedents to $x$ and $w(x)$ are distance-derived softmax weights.

\subsection{Distance Metrics}

Instructions are embedded via $\phi: \mathcal{X} \to \mathbb{R}^d$~\cite{reimers2019sentencebert} and distances computed in a vectorized $(|\mathcal{E}|, |\mathcal{A}|)$ matrix.

Embedding distance:
\[
d_\text{emb}(x, x_j) = 1 - \cos(\phi(x), \phi(x_j))
\]

Structural distance using sentence-level logic graphs (causal/contrastive/conditional edges):
\[
d_\text{struct}(x, x_j) = \mathrm{GED}(G_x, G_{x_j})
\]
where GED (Graph Edit Distance)~\cite{sanfeliu1983ged} measures the minimum cost of edit operations (node/edge insertions, deletions, substitutions) needed to transform one logic graph into another, capturing logical divergence invisible to surface embeddings.

Hybrid distance:
\[
d = \alpha \cdot d_\text{emb} + (1-\alpha) \cdot d_\text{struct}
\]

\subsection{Dynamic Precedent Selection and Weighting}

For each $x \in \mathcal{E}$, select the $k$ nearest precedents:
\[
\mathcal{A}_k(x) = \arg\!\max_k [-d(x, x_j)]
\]

Weight by distance:
\[
w_i(x) = \frac{\exp(-\lambda d(x, a_i))}{\sum_j \exp(-\lambda d(x, a_j))}
\]

At $\lambda=1.0$, distance 0.2 receives $e^{0.4}\!\approx\!1.49\times$ the unnormalised weight of distance 0.6, so nearer precedents dominate the softmax without erasing the contribution of more distant ones.
Both hyperparameters are tunable: $\lambda$ controls the softmax temperature (higher values sharpen the distribution, emphasising the nearest precedents; lower values flatten it), while $k$ determines the neighbourhood size and can be adjusted to match domain characteristics (sparse vs.\ dense regions, noise levels, computational budget).

DA-RAC (Neighbourhood-difficulty) score:
\[
s(x) = \sum_i w_i(x) \cdot llm\_label_i
\]

Static baselines (ablations): \textit{random}, \textit{centroid} (nearest to mean embedding), \textit{diverse} (greedy farthest-point selection).

\subsection{Prompt Construction and Metrics}

\textbf{DA-RAC at inference.}
DA-RAC operates in two parallel registers.
(i)~Each retrieved precedent is rendered into the judge prompt as a labelled few-shot example with its instruction, both candidate responses, and the human-preferred answer; the judge $f_\theta$ then emits a binary preference (0/1) for the target, which is the prediction used for accuracy.
(ii)~The weighted score $s(x) = \sum_i w_i(x)\, llm\_label_i$ aggregates precedents' \emph{llm\_label} field (vanilla-judge agreement with humans on each precedent) into a continuous neighbourhood-difficulty estimate, used as the predicted probability for Expected Calibration Error (ECE)~\cite{guo2017calibration} and Mean Squared Error (Brier).
Precedent selection uses the instruction embedding $\phi(x)$ only; candidate responses and the target's label do not influence retrieval, ruling out trivial label leakage at test time.

Three encoding strategies are designed: \emph{Order} (sort by weight, default), \emph{Repetition} (repeat $\propto w_i$), and \emph{Explicit} (printed scores). Binary output is recorded over a 16-token cutoff. The CoT Rubric baseline uses the five-criterion system prompt with step-by-step reasoning~\cite{liu2023geval}.

\textbf{Metrics:} Mean Squared Error (Brier); Expected Calibration Error (ECE) (10 bin); interpretive misrecognition rate (the fraction of cases in which the judge prefers a rejected artifact).
DA-RAC's weighted score $s(x)$ serves as predicted probability against \emph{llm\_label} (72:28 split).

\section{Experimental Setup}
\label{sec:experimental-setup}
This work uses LLMEval2~\cite{zhang2023llmeval2} as a controlled probe of reference dependence in LLM judging.
This lets us isolate the mechanism that is central to artifact evaluation: LLM judgments change when the supplied precedents are irrelevant, static, or dynamically matched.

\paragraph{Setup.}
LLMEval2: 1,600 samples split in a 70:30 ratio to precedents (577 easy, 223 hard), and 480 test. Instruction embeddings are created using \emph{all-MiniLM-L6-v2}~\cite{reimers2019sentencebert} (384 dimension). For judging, we use \emph{gpt-4o-mini} and \emph{gpt-5.1}; with three runs each (deterministic precedent selection, variance $=$ judge stochasticity). $k=5$, $\lambda=1.0$, strategy $=$ \emph{Order} (sort by weight).
$n{=}50$ subset is used for multi-run judge-stochasticity experiments (as proof of concept and to minimize to inference costs), while full $n{=}480$ set is used for distance-correlation, structural-mismatch, and neighborhood-difficulty analyses.

\section{Probe: Reference Dependence in Judging}
\label{sec:results}

\subsection{Dynamic Precedents Improve Judgment Stability}

Table~\ref{tab:exp1} reports multi-run validated accuracy and Brier scores on 50 evaluation examples.
Both metrics are measured against \emph{human\_label} (for LLMEval2 this is always 0, since it is fully human labeled), so the degenerate always-predict-0 baseline is included for transparency: it achieves 100\%/0.000 by construction, making no inference.
Among methods that perform actual evaluation, DA-RAC achieves 91.3\% $\pm$ 2.5\%, outperforming Vanilla zero-shot by 21\% and CoT Rubric by 17\%.

\begin{table}[t]
\centering
\footnotesize
\caption{Multi-run results (n=50, gpt-4o-mini, 3 runs). Accuracy and Brier vs \texttt{human\_label}$=0$. All variance is judge stochasticity. $^\ddagger$Excluded. $^*$Pre-computed GPT-4.}
\label{tab:exp1}
\begin{tabular}{lccc}
\toprule
\textbf{\textsc{Method}} & \textbf{\textsc{Acc.}} & \textbf{\textsc{Std}} & \textbf{\textsc{Brier}} \\
\midrule
Always-Predict-0$^\ddagger$ & 100 & --- & 0.000 \\
\midrule
\textbf{DA-RAC (ours)} & \textbf{91.3} & \textbf{±2.5} & \textbf{0.087} \\
CoT Rubric & 74.7 & ±0.9 & 0.253 \\
Vanilla (zero-shot) & 70.0 & ±1.6 & 0.300 \\
Static (centroid) & 66.7 & ±5.0 & 0.333 \\
Static (diverse) & 64.7 & ±4.1 & 0.353 \\
Static (random) & 52.0 & ±0.0 & 0.480 \\
\midrule
Dataset LLM (GPT-4)$^*$ & 64.0 & --- & 0.360 \\
\bottomrule
\end{tabular}
\end{table}

\textbf{Reference context changes judgment.}
Static-random precedents achieve 52.0\% $\pm$ 0.0\%, eighteen points below the Vanilla zero-shot baseline (70.0\%). Irrelevant context actively misleads
the judge — adding examples is not automatically beneficial: irrelevant precedents can distort the judge's interpretation. For artifact AI evaluation, this is the central point. The question is not whether a judge has context, but whether it has the \emph{right} context.

\textbf{Interpretive anchoring requires relevance.}
Static precedent strategies plateau at 52--67\%, while DA-RAC's per-example selection reaches 91.3\% $\pm$ 2.5\%.
This supports the design principle that evaluative precedents should be selected relative to the target artifact rather than fixed globally.

\textbf{Rubrics do not replace precedents.}
The CoT rubric baseline is stable (74.7\% $\pm$ 0.9\%) but weaker than DA-RAC.
This suggests explicit criteria alone may not be sufficient for situated evaluation: examples provide concrete interpretive precedents that abstract rubrics may fail to capture.

\begin{figure}[t]
	\centering
	
	\subfloat[Distribution across runs.]{
		\includegraphics[width=\linewidth]{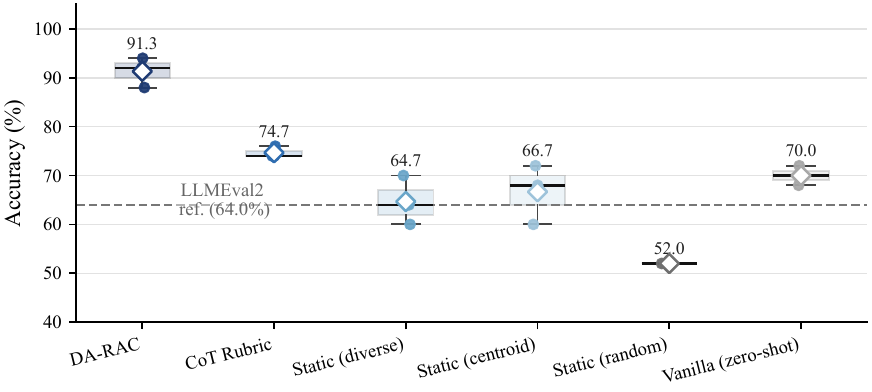}
		\label{fig:multirun_dist}
	}
	
	\vspace{-1mm}
	
	\subfloat[Per-run trajectories.]{
		\includegraphics[width=\linewidth]{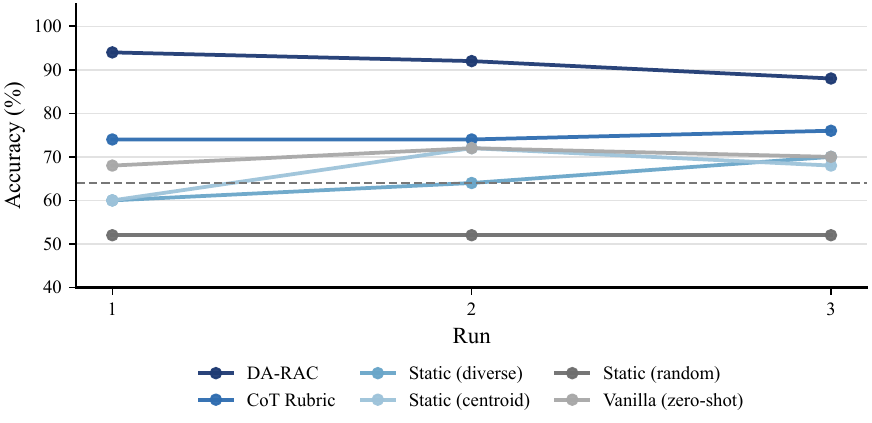}
		\label{fig:multirun_traj}
	}
	
	\caption{
		Multi-run evaluation on LLMEval2 using GPT-4o-mini ($n=50$, $k=5$, three runs).
		DA-RAC achieves the highest mean accuracy ($91.3\%$) with low inter-run variance.
		In Fig.~\ref{fig:multirun_dist}, boxes denote the interquartile range, center lines denote medians, whiskers denote min--max ranges, and diamonds denote means.
		Fig.~\ref{fig:multirun_traj} shows per-run trajectories across methods.
	}
	
	\label{fig:multirun}
\end{figure}

\subsection{Stronger Models Do Not Eliminate Reference Dependence}

Table~\ref{tab:gpt51} reports results with \emph{gpt-5.1}, where DA-RAC achieves 84.7\% ± 1.9\%, with significant gains over Vanilla (+12.0\%), Static-centroid (+16.0\%), and CoT (+20.7\%).

The gap versus Static-diverse narrows to +4.0\% ($p=0.149$), suggesting stronger models reduce but do not eliminate reference dependence.
On gpt-4o-mini, DA-RAC's 91.3\% even exceeds GPT-5.1's vanilla zero-shot at 72.7\%—a smaller judge with better interpretive anchoring outperforms a stronger unanchored one, indicating that scale alone does not remove the sensitivity to precedent selection.
Applying DA-RAC to gpt-5.1 nonetheless improves substantially over gpt-5.1 zero-shot, indicating model capability and precedent selection are complementary rather than substitutes.

\begin{table}[t]
\centering
\footnotesize
\caption{GPT-5.1 results (n=50, 3 runs). DA-RAC shows significant gains over Vanilla ($p=0.001$), centroid ($p<0.001$), CoT ($p<0.001$).}
\label{tab:gpt51}
\begin{tabular}{lcccl}
\toprule
\textbf{\textsc{Method}} & \textbf{\textsc{R1}} & \textbf{\textsc{R2}} & \textbf{\textsc{R3}} & \textbf{\textsc{Mean $\pm$ Std}} \\
\midrule
\textbf{DA-RAC} & 86 & 82 & 86 & \textbf{84.7±1.9} \\
Static (diverse) & 80 & 80 & 82 & 80.7±0.9 \\
Static (random) & 78 & 76 & 74 & 76.0±1.6 \\
Vanilla (zero-shot) & 74 & 72 & 72 & 72.7±0.9 \\
Static (centroid) & 70 & 68 & 68 & 68.7±0.9 \\
CoT Rubric & 64 & 64 & 64 & 64.0±0.0 \\
\midrule
LLMEval2 paper ref & --- & --- & --- & 64.0 \\
\bottomrule
\end{tabular}
\end{table}

\subsection{Distance-Score Correlations and Interpretive Grounding}

\begin{figure}[t]
  \centering
  \includegraphics[width=\linewidth]{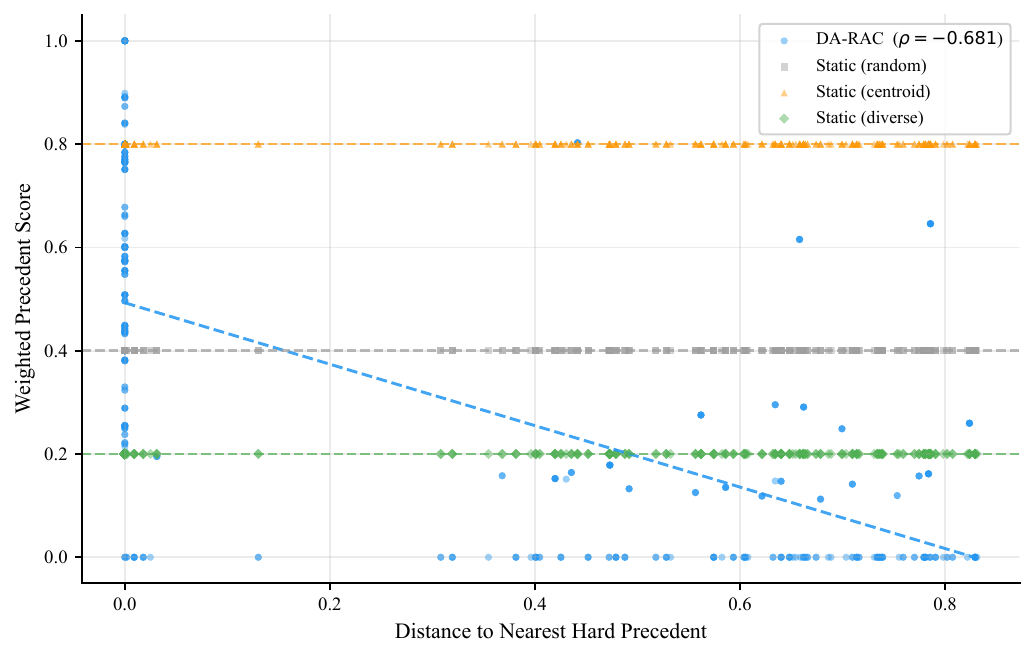}
  \caption{Distance-score correlation (n=480). DA-RAC: $\rho=-0.681$ ($p<0.001$); static methods: $\rho=0.000$. DA-RAC's judgment is grounded in the relevance of its precedent context.}
  \label{fig:distance_score}
\end{figure}

DA-RAC's weighted scores show a strong correlation with precedent distance: $\rho=-0.681$ (hard, $p<0.001$) and $\rho=+0.356$ (easy, $p<0.001$).
Static methods produce constant scores ($\rho=0.000$), confirming DA-RAC's judgment is geometrically grounded in its precedent context rather than detached from it.
Figure~\ref{fig:distance_score} shows DA-RAC's negative correlation with hard-precedent distance; static methods show flat lines.

\subsection{Structural Distance Captures Mismatched Interpretive Form}

\textbf{Structural form matters for artifact evaluation.}
Hybrid distance ($\alpha=0.5$) detects 91 examples (19\%) that are surface-similar but logically divergent---cases where embedding-based retrieval selects misleading precedents. Static methods, by construction, rely on surface similarity, thus fail on such examples (52--67\% accuracy, per Table 1). Cultural artifacts often share vocabulary but differ in argument, genre, contrastive structure, causality, or conditional logic. DA-RAC's hybrid distance lets the system retrieve precedents that match not just topic but interpretive form.

\textbf{When structural distance helps.}
Form-level distance is most beneficial for evaluations involving complex reasoning: contrastive comparisons (``A but not B''), causal chains (``if X then Y''), and conditional logic (``under Z, prefer A'').
For tasks turning on surface similarity (style or topical match), embedding distance alone suffices (the 81\% where $d_\text{emb}$ and $d_\text{struct}$ agree).
The hybrid setting ($\alpha=0.5$) provides robustness: GED corrects the 19\% form-mismatched edge cases while embeddings handle the majority efficiently.

\subsection{Neighbourhood Difficulty as Contestability}

Examples are partitioned by $r = d_\text{hard}/(d_\text{easy}+d_\text{hard})$ into confident-easy ($r>0.60$, n=231), ambiguous ($0.40 \le r \le 0.60$, n=50), and confident-hard ($r<0.40$, n=199) zones.

DA-RAC achieves ECE 0.084 and Brier 0.173, while
static-centroid is severely miscalibrated in the easy zone (ECE 0.666), where its fixed prediction over-states difficulty.
In the ambiguous zone, where nearest precedents are split between easy and hard exemplars, DA-RAC reaches 92.0\% judge accuracy versus static-centroid's 52.0\% (near chance).
For cultural AI design, the ambiguous zone is the use-case that should be flagged for human review: the neighbourhood-difficulty score makes such cases inspectable rather than silently judged.

\section{From Calibration to Reliable Evaluation Design}
\label{sec:design}

DA-RAC can be read not only as a calibration method but as a design pattern for culturally situated evaluation.
A cultural evaluation workflow built on DA-RAC has four components, summarised in Table~\ref{tab:interpretive_design}.

\paragraph{Curated precedent pools.}
Instead of drawing precedents from arbitrary benchmark examples, a reliable AI evaluation system should use precedent pools curated by the relevant communities, domain experts, artists, critics, educators, or practitioners~\cite{costanza2020design}.
These precedents encode what counts as successful performance for a given genre, audience, or value.

\paragraph{Inspectable interpretive lineage.}
For each judgment, DA-RAC exposes the retrieved precedents and their weights.
This makes the judge's effective context visible: reviewers can inspect which precedents shaped the verdict and contest whether those precedents are appropriate.

\paragraph{Contestability through neighbourhood difficulty.}
High nearest-precedent distance, mixed neighborhood labels, or disagreement between semantic and structural distance should not be hidden.
These cases signal interpretive uncertainty.
In cultural domains, uncertainty may not be noise but rather indicate genuine ambiguity, plural readings, or need for community review. So, a structure distance will help — such as the GED distance method proposed in this paper.

\paragraph{Human agency by design.}
DA-RAC should not be assumed to automate cultural authority~\cite{suchman1987plans}.
Instead, it can support human-AI ensembles: offering relevant precedents and a provisional judgment, while humans retain authority over ambiguous, novel, or contested cases.

\begin{table*}[t]
\centering
\footnotesize
\caption{DA-RAC as an interpretive technology for AI evaluation.}
\label{tab:interpretive_design}
\begin{tabular}{p{0.20\linewidth}p{0.27\linewidth}p{0.34\linewidth}}
\toprule
\textbf{\textsc{DA-RAC component}} & \textbf{\textsc{Artifact interpretation}} & \textbf{\textsc{Design function}} \\
\midrule
Precedent pool          & Curated precedents      & Encodes local criteria \\
Distance retrieval      & Context matching        & Selects relevant examples \\
Softmax weights         & Precedent strength      & Shows influence on judgment \\
Structural distance     & Interpretive form       & Detects genre/logic mismatch \\
Neighbourhood difficulty & Contestability         & Flags cases for review \\
Logged precedents       & Interpretive lineage    & Enables human inspection \\
\bottomrule
\end{tabular}
\end{table*}

\paragraph{Visualizaton: community-facing cultural explanation.}
Consider an AI system that generates short explanatory labels for a community archive or local museum collection.
A generic evaluator might reward fluent, polished, encyclopedic prose.
But the relevant cultural value may be different: preserving local terminology, acknowledging contested histories, avoiding institutional flattening, or making space for community memory.
In this setting, DA-RAC's precedent pool would consist of community-curated examples of successful and unsuccessful labels.
For each generated label, the judge would retrieve nearby precedents, expose them to reviewers, and flag cases whose nearest precedents are distant or contested.
The positive outcome is not that the AI learns to ``do culture'' (or tasks) autonomously, but that evaluation becomes more context-sensitive and reviewable by the people whose fields / materials are at stake.

\section{Discussion}
\label{sec:discussion}

\paragraph{Evaluation as situated interpretation.}
The main takeaway of the work is not a simple accuracy improvement of LLM-as-judges, but rather the observation that LLM-evaluations are highly sensitive to examples used to frame a task. For Generative AI producing or evaluating cultural artifacts, this sensitivity should be paramount, and as such should not be evaluated against a universal criteria alone. Rather they are to be interpreted through precedents, genres, audiences, and values~\cite{gadamer1975truth,hall1980encoding}.

\paragraph{Value add: contextual sensitivity.}
Much work on AI evaluation focuses on avoiding harms such as bias, misinformation, or moral violation. These goals are necessary but incomplete.
A positive account of cultural evaluation should ask what success looks like. We propose contextual sensitivity as one such value: an evaluator should ground its judgment in relevant precedents, expose those precedents, and recognise when a case exceeds its available context.

\paragraph{Why distance matters.}
DA-RAC, by combining both structural and surface (semantic) distances, operationalizes this idea by making precedent selection distance-aware. Static or random references can impose an less-accurate interpretive frame. Dynamic retrieval makes the frame local to the target. Structural distance further matters because two artifacts can be topically similar while differing in argument, genre, contrast, causality, or conditional structure.

\paragraph{Contestability rather than automation.}
The goal of the work is not to make LLM judges into final arbiters of cultural value, but to make their judgments more contestable.
By logging retrieved precedents and exposing neighbourhood difficulty, DA-RAC creates opportunities for human reviewers to ask: were these the right examples?
Do they reflect the relevant community?
Is this case genuinely ambiguous?
Should another interpretive tradition be represented?

\paragraph{Limitations.}
Our empirical results are based on LLM-judge benchmarks rather than a community-specific cultural dataset.
We therefore present them as a technical probe of reference dependence, not as evidence that DA-RAC ``solves'' cultural/artifact evaluation.
In this benchmark implementation, precedent retrieval uses instruction embeddings only, to avoid target-label leakage; in AI deployments, the representation should also incorporate non-label-bearing contextual metadata such as genre, audience, community, medium, or artifact descriptors.
Future work may construct community-curated precedent pools and evaluate DA-RAC in domains such as creative-writing feedback, museum labels, culturally specific health communication, public-memory projects, and multilingual civic explanation.

\section{Conclusion}
\label{sec:conclusion}

LLM-as-a-judge is a prime candidate for treatment as an interpretive technology: its judgments about generated artifacts depend on relevant precedents, genre, audience, and value rather than on context-free criteria.
DA-RAC operationalises this view through distance-aware interpretive anchoring---retrieving semantically and structurally proximate precedents, weighting them by distance, and exposing neighbourhood difficulty as a signal for human review.

Our benchmark results show that less relevant references can substantially distort LLM judgments, while distance-aware anchoring improves calibration relative to zero-shot, rubric-based, and static-anchor baselines. We interpret these results as a probe of a broader AI design principle: evaluation should be context-sensitive, inspectable, and contestable.

A positive vision for such evaluation should not necessarily aim to automate cultural authority.
It should build systems that help humans see how judgments are made, when context is missing, and where interpretation should remain open.
DA-RAC offers one mechanism toward this goal.

\section*{Impact Statement}
This work aims to support culturally situated evaluation, not to automate cultural authority.
LLM judges can flatten differences across genres, communities, and interpretive traditions, especially when prompted with irrelevant or dominant-culture precedents.
DA-RAC may reduce this risk by making reference selection distance-aware, inspectable, and contestable.
However, precedent pools can encode canon bias, institutional preferences, exclusionary norms, or dominant cultural assumptions.
In creative and cultural domains, DA-RAC should therefore be used to support artists, critics, communities, educators, and domain experts, not to replace them.
We recommend community-curated precedent pools, logging retrieved precedents, reporting performance across cultural domains, and routing high-distance or high-disagreement cases to human review.
The value pursued here is contextual sensitivity with human agency: AI evaluation should help humans interpret and contest cultural judgments rather than silently automate them.


\bibliographystyle{icml2026}

\end{document}